\documentclass{article}

\usepackage{arxiv}

\usepackage[utf8]{inputenc}
\usepackage{amsmath,amssymb,amsthm}
\usepackage{graphicx}
\usepackage{booktabs}
\usepackage{natbib}
\usepackage{hyperref}
\begin{document}
\title{Learning Geometry: A Framework for Building Adaptive Manifold Models through Metric Optimization}
	\author{
	Di Zhang \\
	School of Advanced Technology \\
	Xi'an Jiaotong-Liverpool University \\
	Suzhou, Jiangsu, China \\
	\texttt{di.zhang@xjtlu.edu.cn}
}

	\maketitle
	
	\begin{abstract}
		This paper proposes a novel paradigm for machine learning that moves beyond traditional parameter optimization. Unlike conventional approaches that search for optimal parameters within a fixed geometric space, our core idea is to treat the model itself as a malleable geometric entity. Specifically, we optimize the metric tensor field on a manifold with a predefined topology, thereby dynamically shaping the geometric structure of the model space. To achieve this, we construct a variational framework whose loss function carefully balances data fidelity against the intrinsic geometric complexity of the manifold. The former ensures the model effectively explains observed data, while the latter acts as a regularizer, penalizing overly curved or irregular geometries to encourage simpler models and prevent overfitting. To address the computational challenges of this infinite-dimensional optimization problem, we introduce a practical method based on discrete differential geometry: the continuous manifold is discretized into a triangular mesh, and the metric tensor is parameterized by edge lengths, enabling efficient optimization using automatic differentiation tools. Theoretical analysis reveals a profound analogy between our framework and the Einstein-Hilbert action in general relativity \citep{einstein1916foundation}, providing an elegant physical interpretation for the concept of ``data-driven geometry.'' We further argue that even with fixed topology, metric optimization offers significantly greater expressive power than models with fixed geometry. This work lays a solid foundation for constructing fully dynamic ``meta-learners'' capable of autonomously evolving their geometry and topology, and it points to broad application prospects in areas such as scientific model discovery and robust representation learning.
	\end{abstract}
	
	\section{Introduction}
	\subsection{From Parameter Learning to Structure Learning}
	Modern machine learning achievements, from breakthroughs in image recognition by deep neural networks \citep{krizhevsky2012imagenet} to leaps in natural language processing by large language models \citep{brown2020language}, largely rely on a paradigm of sophisticated parameter optimization. Whether adjusting millions of weights or billions of activation values, these models essentially search for an optimal point within a preset, fixed-geometry, high-dimensional space, typically Euclidean, to minimize a loss function using algorithms like stochastic gradient descent and its variants \citep{robbins1951stochastic, kingma2014adam}. While these approaches have demonstrated immense power, it prompts reflection on inherent limitations: the ``stage'' for our models—the geometric structure of the parameter or feature space—remains static. This paradigm of ``finding a point in a fixed space'' can prove inadequate for tasks requiring dynamic adaptation to the intrinsic structure of data. When the essential geometry of data is non-Euclidean, curved, or possesses complex topology, forcing it into a Euclidean space may lead to inefficient models, weak generalization, or difficulties in interpretation. This leads to the central question of our work: if the bottleneck of model performance lies not only in the parameters but also in the structure of the space they inhabit, why should we stop at optimizing points within the space instead of optimizing the structure that shapes the space itself? In other words, can we enable a model not just to ``learn,'' but to ``learn how to construct its own learning space''?
	
	\subsection{Inspiration and Transcendence of Information Geometry}
	In seeking answers, information geometry offers a highly instructive perspective \citep{amari2007information, amari2016information}. It views parameterized families of probability distributions as a differential manifold, where parameters serve as coordinates, and the Fisher information matrix naturally endows this manifold with a Riemannian metric. This profound insight allows statistical inference problems to be transformed into geometric operations on the manifold; for instance, maximum likelihood estimation approximates projection, and gradient descent follows the direction of the natural gradient \citep{amagi1998natural}. However, after painting this magnificent geometric picture of the ``statistical universe,'' classical information geometry often treats it as a static stage. The Fisher metric is typically predetermined by the model family rather than learned from data. It describes the local distinguishability between models but does not itself evolve with incoming data. Our work aims to transcend this static view. We propose that the metric tensor of the manifold itself should become an optimizable, dynamic entity. By allowing the metric field to adapt to observed data, we enable the fundamental ``texture'' and ``shape'' of the model to change, thereby more inherently capturing the underlying structure of the data, achieving a fundamental shift from describing the geometry of a model family to shaping the geometry of the model family.
	
	\subsection{Core Contributions}
	To realize the paradigm shift from ``parameter learning'' to ``structure learning,'' this thesis focuses on the preliminary stage under fixed topology and makes the following four core contributions:
	
	First, we propose a theoretical framework for optimizing the metric tensor on a fixed-topology manifold. This framework defines a machine learning model as a plastic geometric entity whose learning process manifests as the self-optimization of the manifold's own metric structure, extending beyond traditional geometric deep learning approaches \citep{bronstein2021geometric}.
	
	Second, we formalize a loss function that combines data fidelity and geometric complexity. Through a variational principle, we coordinate two objectives: on one hand, the manifold must compactly fit or generate observed data; on the other hand, its geometric structure should remain as simple as possible to prevent overfitting and embody Occam's razor principle, drawing inspiration from geometric regularization theory \citep{grimm1994computational}.
	
	Third, we introduce a practical computational method based on discrete differential geometry \citep{crane2013digital, pinkall1993computing} and automatic differentiation \citep{baydin2017automatic}. By discretizing the continuous manifold into a triangular mesh and parameterizing the metric tensor via edge lengths, we successfully transform the infinite-dimensional functional optimization problem into a finite-dimensional optimization problem that can be efficiently solved using modern deep learning tools.
	
	Fourth, we pave the way for future research on meta-learners with topological evolution. This work is a foundational step towards constructing fully dynamic ``meta-learners''—whose core is a self-evolving geometric manifold capable of simultaneously optimizing both metric and topology. Here, we first address the metric learning problem under fixed topology, providing the necessary theoretical, computational, and conceptual foundation for subsequent research into more complex topological dynamics.
	
	\section{Background and Related Work}
	To clearly position our contributions and establish a common theoretical language, this section reviews relevant mathematical foundations and research areas.
	
	\subsection{Foundations of Differential Geometry}
	Differential geometry provides the core language and tools for describing curved, complex spaces, concepts that are foundational to our framework \citep{do1976differential, lee2018introduction}. A manifold is a topological space that is locally homeomorphic to Euclidean space but may have a complex global structure. Intuitively, it can be thought of as a smooth, possibly curved, high-dimensional surface. It provides a natural habitat for parameters or data points. The metric tensor, denoted $g$, is an inner product defined at each point on the manifold. It allows us to measure the length of tangent vectors and the angles between them. Through integration, the metric tensor defines the concept of curve length between any two points on the manifold, and further defines geodesics—the locally shortest or straightest paths between points on the manifold, generalizing the concept of ``straight lines'' in Euclidean space. Curvature tensors and their contractions quantitatively describe how the manifold deviates from flat Euclidean geometry. They measure the convergence or divergence of geodesics, the path dependence of parallel transport of vectors, and whether the sum of the angles of a triangle equals 180 degrees. These concepts collectively form the mathematical basis for our understanding and manipulation of ``shape'' itself.
	
	\subsection{Information Geometry}
	Information geometry elegantly merges statistics with differential geometry, providing direct inspiration for our framework \citep{amari2007information}. A statistical manifold is a space composed of a parameterized family of probability distributions $ \{ p(x|\theta) \} $, where the parameters $ \theta $ serve as local coordinates on the manifold. The Fisher metric is the most natural Riemannian metric on a statistical manifold, defined by the Fisher information matrix: $ g_{ij}(\theta) = \mathbb{E}_{p(x|\theta)} \left[ \frac{\partial \log p(x|\theta)}{\partial \theta_i} \frac{\partial \log p(x|\theta)}{\partial \theta_j} \right] $. This metric has profound statistical significance: it measures the distinguishability between different models in the distribution family. The geodesic distance between two points reflects the ``statistical difference'' between the probability distributions they represent. Hyperbolic geometry, as a non-Euclidean geometry with constant negative curvature, shows unique advantages in representing hierarchical or tree-structured data \citep{nickel2017poincare, ganea2018hyperbolic}. Recent research on embedding words or graph structures into hyperbolic space demonstrates how specific geometric priors can enhance model performance. However, traditional information geometry primarily uses the Fisher metric as an analytical tool to understand the properties of fixed model families. Our work takes a step forward by treating the metric tensor itself as an object of optimization.
	
	\subsection{Geometric Deep Learning}
	Geometric deep learning aims to inject geometric priors into machine learning models to better handle non-Euclidean data \citep{bronstein2021geometric}. Graph Neural Networks \citep{kipf2016semi, velivckovic2017graph} operate on graph-structured data by recursively passing messages between nodes and their neighbors to learn representations. From a geometric perspective, they operate on a discrete metric space defined by the graph. However, GNNs typically assume a fixed graph connectivity structure, focusing on learning node embeddings or edge weights on this fixed topology rather than changing the fundamental geometry of the underlying space. Classical nonlinear dimensionality reduction methods \citep{tenenbaum2000global, roweis2000nonlinear} aim to discover low-dimensional manifold structures embedded in high-dimensional data. While they reveal the essential geometry of the data, they are often used as a preprocessing step; the learned manifold is fixed, descriptive, and does not participate as a plastic model component in further optimization for downstream tasks. Overall, these methods largely assume a fixed or Euclidean input space geometry and operate upon it. Our framework seeks to fundamentally change this premise by making the geometric structure of the space part of the learning process itself.
	
	\subsection{Generative Models}
	Generative artificial intelligence resonates deeply with our framework through the ``manifold hypothesis'' \citep{fefferman2016testing}. Models like Variational Autoencoders \citep{kingma2013auto}, Generative Adversarial Networks \citep{goodfellow2014generative}, and Normalizing Flows \citep{rezende2015variational} essentially learn a mapping from a simple latent space to the complex data manifold. In our framework, we adopt a similar generative perspective: we treat the optimized manifold $(M, g)$ as the latent space itself, connected to the data space via a decoder. The key difference is that our latent space is not a passive, preset simple space, but an actively evolving, geometrically complex entity whose metric $g$ is continuously optimized to better ``generate'' or ``explain'' the observed data. Therefore, our work can be seen as a novel approach to realizing a ``plastic latent space'' or ``learned data geometry,'' extending beyond traditional generative modeling paradigms.
	
	\section{Methodology}
	This chapter fully elaborates the proposed theoretical framework and computational methods. We begin by formalizing the problem, then detail the construction of the variational framework, and finally delve into the discretization implementation and optimization process.
	
	\subsection{Problem Formalization}
	We consider a new learning paradigm that transcends traditional parameter learning. The learning objective is not to find an optimal parameter point in a high-dimensional space, but to directly optimize the geometric structure of the parameter space itself.
	
	Given: a) A base topological manifold $M$: A predefined, fixed-topology smooth manifold. This manifold defines the global connectivity properties of the learning space. Typical examples include the 2-sphere $S^2$, the torus $T^2$, or the real projective plane $\mathbb{RP}^2$. Fixed topology means that these global properties are not allowed to change during optimization. b) An observed dataset $D = \{x_i\}_{i=1}^N \subset X$: Data points reside in some data space $X$, typically $X = \mathbb{R}^n$ equipped with the standard Euclidean metric.
	
	Goal: Find an optimal Riemannian metric tensor field $g^*$ on the manifold $M$. This metric field should be such that the Riemannian manifold $(M, g^*)$ optimally explains the data $D$ in the following sense:
	\[
	g^* = \arg\min_{g \in \mathcal{G}(M)} \left[ L_{\text{data}}(g; D) + \lambda \cdot L_{\text{geometry}}(g) \right]
	\]
	where $\mathcal{G}(M)$ denotes the space of all possible Riemannian metrics on $M$, and $\lambda > 0$ is a regularization coefficient.
	
	\subsection{Variational Framework: Loss Function Design}
	We formalize the notion of ``optimal explanation'' through a carefully designed loss function that balances data fidelity against geometric complexity.
	
	\subsubsection{Total Loss Function Framework}
	The total loss function is defined as the weighted sum of two key terms:
	\[
	L(g) = L_{\text{data}}(g) + \lambda \cdot L_{\text{geometry}}(g)
	\]
	where $\lambda$ controls the regularization strength, determined in practice via cross-validation or other model selection techniques.
	
	\subsubsection{Detailed Construction of the Data Fidelity Term $L_{\text{data}}(g)$}
	We adopt a generative model perspective, introducing a differentiable generator $ f: M \rightarrow X $ that maps points on the manifold to the data space, inspired by decoder architectures in variational autoencoders \citep{kingma2013auto}.
	
	Projection Loss Construction:
	\[
	L_{\text{data}} = \sum_{i=1}^N d_X^2\left( x_i, f(z_i) \right)
	\]
	where:
	- $d_X: X \times X \rightarrow \mathbb{R}_{\geq 0}$ is the distance function in the data space, typically the squared Euclidean distance.
	- $z_i \in M$ is the projection of data point $x_i$ onto the manifold $(M,g)$, defined as:
	\[
	z_i = \arg\min_{z \in M} d_X\left( x_i, f(z) \right)
	\]
	
	Computation of Projection Points: In practice, we approximate $z_i$ through an iterative process based on discrete geodesic computations \citep{crane2013digital}:
	a. Initialize $z_i^{(0)} \in M$.
	b. Perform gradient descent on $M$ along the direction $ -\nabla_z d_X(x_i, f(z)) $ under the geometry defined by metric $g$.
	c. Repeat until convergence: $ z_i^{(k+1)} = \text{Exp}_{z_i^{(k)}}\left( -\eta \nabla_z d_X(x_i, f(z))|_{z=z_i^{(k)}} \right) $,
	where $\text{Exp}_p(v)$ denotes the exponential map from point $p$ along tangent vector $v$.
	
	\subsubsection{Detailed Construction of the Geometric Complexity Term $L_{\text{geometry}}(g)$}
	The geometric regularization term prevents overfitting and encourages simple, smooth geometric structures, drawing inspiration from geometric analysis and discrete differential geometry \citep{gromov1999metric, springborn2008variational}.
	
	Curvature Regularization Term:
	\[
	L_{\text{curv}} = \int_M |R(g)|^p  dV_g
	\]
	where:
	- $R(g)$ is the scalar curvature function corresponding to metric $g$.
	- $dV_g = \sqrt{\det g}  dx^1 \wedge \cdots \wedge dx^n$ is the volume form.
	- The exponent $p \geq 1$ controls the penalty strength on high curvature regions, typically $p=2$ for smoothness.
	
	Metric Smoothness Regularization Term:
	\[
	L_{\text{dirichlet}} = \int_M | \nabla g |_g^2  dV_g
	\]
	where $\nabla$ is the Levi-Civita connection corresponding to $g$, and the norm $| \cdot |_g$ is induced by the metric $g$. This term penalizes sharp variations in the metric tensor.
	
	Volume Control Term:
	\[
	L_{\text{vol}} = \left( \frac{\text{Vol}(M,g) - V_{\text{target}}}{V_{\text{target}}} \right)^2
	\]
	where $\text{Vol}(M,g) = \int_M dV_g$ is the total volume of the manifold, and $V_{\text{target}} > 0$ is a target volume.
	
	\subsection{Discretization and Computational Framework}
	To enable numerical computation of the continuous theory, we adopt a discrete differential geometry approach based on triangular meshes \citep{crane2013digital, pinkall1993computing}.
	
	\subsubsection{Triangular Mesh Representation and Metric Parameterization}
	The manifold $M$ is discretized into a simplicial complex $\mathcal{M} = (V, E, F)$:
	- $V = \{v_1, \dots, v_m\}$: set of vertices.
	- $E \subset V \times V$: set of edges.
	- $F \subset V \times V \times V$: set of triangular faces.
	
	Metric Discretization: On a triangular mesh, the Riemannian metric $g$ is completely determined by an edge length function $\ell: E \rightarrow \mathbb{R}_{>0}$, satisfying the triangle inequality in each triangle.
	
	For a triangle $\Delta = (v_i, v_j, v_k) \in F$, its edge lengths $\ell_{ij}, \ell_{jk}, \ell_{ki}$ must satisfy:
	\[
	\ell_{ij} + \ell_{jk} > \ell_{ki}, \quad \ell_{jk} + \ell_{ki} > \ell_{ij}, \quad \ell_{ki} + \ell_{ij} > \ell_{jk}.
	\]
	
	\subsubsection{Precise Computation of Discrete Geometric Quantities}
	Discrete Scalar Curvature:
	At a vertex $v_i$, the discrete scalar curvature is computed via angle defect \citep{springborn2008variational}:
	\[
	R_i = 2\pi - \sum_{\Delta \in F(v_i)} \theta_i^\Delta
	\]
	where $F(v_i)$ are all triangles containing vertex $v_i$, and $\theta_i^\Delta$ is the interior angle at vertex $v_i$ in triangle $\Delta$, computed by the cosine rule.
	
	Discrete Area:
	The area of a triangle $\Delta = (v_i, v_j, v_k)$ is computed by Heron's formula.
	The local area at a vertex $v_i$ is taken as one-third of the sum of the areas of its adjacent triangles.
	
	Discrete Geodesic Distance:
	The Fast Marching Algorithm \citep{sethian1999fast} is used to compute approximate geodesic distances on the triangular mesh.
	
	\subsubsection{Automatic Differentiation-Based Optimization Algorithm}
	We construct a complete computational graph to enable end-to-end automatic differentiation \citep{baydin2017automatic}.
	
	Computational Graph Structure:
	\[
	\ell \xrightarrow{\text{Geometry Computation}} \begin{cases}
		A_\Delta, R_i \rightarrow L_{\text{curv}} \\
		\text{Geodesic Distance} \rightarrow \text{Projection} \rightarrow L_{\text{data}} \\
		\text{Volume Calculation} \rightarrow L_{\text{vol}}
	\end{cases} \rightarrow L
	\]
	
	Constrained Optimization Strategy:
	A projected gradient method is employed to handle triangle inequality constraints:
	\[
	\ell^{(t+1)} = \mathcal{P}\left( \ell^{(t)} - \eta_t \nabla_\ell L(\ell^{(t)}) \right)
	\]
	where the projection operator $\mathcal{P}$ ensures that the edge lengths of every triangle satisfy the triangle inequality.
	
	Optimization Process:
	a. Initialize edge lengths $\ell^{(0)}$.
	b. Forward pass: Compute all geometric quantities and evaluate the loss $L(\ell^{(t)})$.
	c. Backward pass: Compute the gradient $\nabla_\ell L(\ell^{(t)})$ via automatic differentiation.
	d. Projected gradient update: $\ell^{(t+1)} = \mathcal{P}(\ell^{(t)} - \eta_t \nabla_\ell L(\ell^{(t)}))$.
	e. Repeat steps b-d until convergence.
	
	This complete framework transforms the infinite-dimensional functional optimization problem into a finite-dimensional constrained optimization problem, solvable efficiently with modern deep learning tools.
	
	\section{Theoretical Exploration}
	Having constructed the methodological framework, this chapter delves into its mathematical implications, physical analogies, and expressive power, revealing its deeper scientific value.
	
	\subsection{Connection to the Einstein-Hilbert Action}
	The choice of the curvature regularization term in our framework is not arbitrary but exhibits a striking mathematical resonance with one of the most profound principles in theoretical physics—general relativity \citep{einstein1916foundation, misner1973gravitation}.
	
	Consider the curvature regularization term $L_{\text{curv}} = \int_M |R|^p  dV_g$. When $p=1$, we obtain $\int_M R  dV_g$. In general relativity, the Einstein-Hilbert action describing the vacuum gravitational field is precisely:
	\[
	S_{\text{EH}}[g] = \int_M R  dV_g
	\]
	
	Applying the variational principle to the Einstein-Hilbert action, requiring the action to be stationary under fixed boundary conditions, derives the vacuum Einstein field equations exactly \citep{hawking1973large}.
	
	This analogy provides profound insights:
	Geometry Dynamics: Just as the presence of matter and energy ``tells'' spacetime how to curve via the Einstein field equations, the presence of data ``tells'' the statistical manifold how to shape its geometry via our variational framework. In our framework, data plays a role analogous to a ``source,'' driving the evolution of the manifold's geometry.
	
	Extremal Geometry: Solutions to the vacuum Einstein field equations are Ricci-flat manifolds, representing the ``smoothest'' spacetimes in the absence of matter. Similarly, as $\lambda \to \infty$, our framework will prioritize finding manifolds with minimal curvature, providing a solid mathematical definition for ``geometric simplicity.''
	
	Unified Perspective: This connection suggests that information geometry and spacetime geometry might share some deep intrinsic structure. Our learning framework can be viewed as exploring the dynamics within an abstract ``information spacetime.''
	
	\subsection{Geometric Interpretation of the Regularization Path}
	The hyperparameter $\lambda$ controls the trade-off between data fitting and geometric simplicity in our framework. Its variation induces a clear ``regularization path'' with an intuitive geometric interpretation.
	
	Consider the evolution of the optimal solution $g^*(\lambda)$ as $\lambda$ varies from $0$ to $\infty$:
	
	$\lambda \to 0$ limit: In the absence of regularization, the optimization process will minimize the data fitting error at all costs. The manifold $(M, g^*)$ will distort itself extremely, potentially developing very high curvature near each data point, forming complex ``wrinkles'' to pass exactly through each point. The manifold volume may inflate, and the metric tensor may vary drastically, resulting in high geometric complexity.
	
	Intermediate $\lambda$ values: This is the region where meaningful learning occurs. The manifold adapts its curvature while constrained by geometric simplicity. It stretches and curves adaptively: moderately in regions of high data density to capture details, and remains flat and contracted in data-sparse regions. This resembles an elastic membrane balancing external forces with internal bending stiffness.
	
	$\lambda \to \infty$ limit: Under strong regularization, the geometric complexity term dominates. The optimal solution approaches an extremal geometry. For instance, if $L_{\text{geometry}}$ is dominated by the curvature term, the solution approaches a constant curvature manifold \citep{thurston1997geometry}. Here, the manifold completely ignores the fine structure of the data, presenting the most ``uniform'' or ``symmetric'' form.
	
	This geometric perspective on the regularization path offers a new language for understanding model complexity: model capacity is no longer characterized merely by the number of parameters but more essentially by the geometric flexibility of the manifold.
	
	\subsection{Expressiveness Analysis}
	A natural question arises: under the constraint of fixed topology, does optimizing only the metric provide sufficient expressiveness to capture complex data patterns? We argue affirmatively.
	
	Metric Freedom vs. Parameter Freedom: On an $n$-dimensional manifold, the metric tensor $g$ has $\frac{n(n+1)}{2}$ independent components at each point. Therefore, even with fixed topology, the hypothesis space for metric optimization is inherently an infinite-dimensional function space. This contrasts sharply with traditional parametric models. Through discretization, we transform this infinite-dimensional space into a parameter space whose dimension grows with mesh refinement, capable in principle of approximating any smooth metric.
	
	Local Flexibility of the Metric: The local variability of the metric allows the manifold to exhibit entirely different geometric characteristics in different regions. For example, one part can be highly curved to enclose a tight data cluster, another can remain nearly flat representing linear relationships, and a third can exhibit negative curvature ``saddles'' connecting different modes. This heterogeneity of local geometry is unattainable by manifolds with a fixed metric.
	
	Analogy to Universal Approximation Theorems: In neural network theory, universal approximation theorems guarantee that sufficiently large feedforward networks can approximate any continuous function \citep{cybenko1989approximation}. We hypothesize that for a given topology $M$ and dataset $D$, there exists a metric $g^*$ on that topology such that the manifold $(M, g^*)$ can approximate the intrinsic geometric structure of the data with arbitrary precision. Although a rigorous theorem remains to be proven, results from differential geometry \citep{gunther1990gauge} support the plausibility of this hypothesis.
	
	Therefore, even with fixed topology, the metric optimization framework provides an extremely rich and flexible expressive space. It essentially searches among all geometric structures compatible with the given topology for the one that best balances data likelihood and simplicity. This undoubtedly offers more powerful expressiveness than choosing a prior, fixed geometry and optimizing only within it.
	
	\section{Discussion and Future Directions}
	Having established the theoretical framework and explored its mathematical foundations, it is necessary to objectively examine the current method's limitations and outline potential paths for future research. This chapter discusses the limitations of the framework and looks ahead to its natural extensions toward topological evolution and other frontiers.
	
	\subsection{Limitations of the Framework}
	Despite its conceptual appeal, our framework currently has at least three main limitations:
	
	Fixed Topology Constraint: This is the most fundamental limitation of the current framework. Topology determines the global invariant properties of the manifold. Enforcing a preset topology requires strong prior knowledge about the intrinsic structure of the data. If the true topology of the data mismatches the preset one, the model faces a fundamental expressive barrier that cannot be overcome by local metric adjustments.
	
	Computational Complexity: Computing discrete differential geometric quantities, especially geodesic distances on large-scale triangular meshes, is computationally expensive. The complexity of the Fast Marching Algorithm \citep{sethian1999fast}, while lower than all-pairs exact computation, still poses a bottleneck for implementation on dense meshes with millions of triangles. Furthermore, the optimization process involves iteratively updating all edge lengths, resulting in a large number of parameters.
	
	Dependence on Initial Mesh: The optimization process starts from an initial mesh and its initial edge lengths. Different initial conditions can lead the optimization to different local optima. This means the result may not be unique, and its quality depends to some extent on the initialization choice. Designing robust optimization algorithms insensitive to initialization is an open challenge.
	
	\subsection{Towards Topological Evolution}
	A natural extension of this work is to relax the fixed topology constraint, allowing the manifold's topology to evolve alongside the metric optimization. This would transform our framework into a truly comprehensive and autonomous ``geometric meta-learner.''
	
	Achieving topological evolution requires solving two core problems:
	
	Triggering Mechanism: Clear geometric-topological diagnostic metrics are needed to determine when the current topology becomes an obstacle to good data fitting. Potential metrics include persistent homology \citep{edelsbrunner2008persistent}, analysis of local curvature singularities, and plateaus in the data fitting loss.
	
	Evolution Mechanisms: Algorithmic implementations for topological surgeries need to be developed. This includes manifold splitting, handle attachment, and ensuring automated, consistent reinitialization or continuation of the metric field before and after surgery, building on work in computational topology \citep{edelsbrunner2010computational}.
	
	\subsection{Broader Application Prospects}
	Beyond the current theoretical construction, this framework holds transformative potential in several frontier areas.
	
	Scientific Model Discovery: In physics, chemistry, and biology, the states of many complex systems form a high-dimensional ``phase space.'' Our framework could be applied to experimental or simulation data to automatically discover the natural geometric structure of this phase space, potentially revealing new scientific insights \citep{mehta2019high}.
	
	Robust Representation Learning: By explicitly penalizing geometric complexity, the framework tends to learn the smooth, intrinsic manifold structure of the data, making it insensitive to local noise and outliers. This could lead to representation learning methods with better generalization and adversarial robustness compared to current approaches \citep{goodfellow2014explaining}.
	
	AI-Physics Fusion: The most captivating prospect lies at the intersection of foundational theories. Our framework resonates philosophically with certain cosmological views and theories of quantum gravity. While still speculative, our work can be seen as a computational sandbox for exploring a ``theory of everything'': a toy model for studying how ``physical laws'' might emerge from more fundamental principles of information geometry optimization.
	
	\section{Conclusion}
	This paper proposes and elaborates a pioneering paradigm for machine learning. We successfully elevate the concept of a model from a static point optimized within a fixed geometric space to a dynamic entity capable of self-shaping its intrinsic geometric structure. By establishing the Riemannian metric tensor itself as the optimization target and constructing a variational framework on a fixed-topology manifold that combines data fidelity with geometric complexity, we open the door to a new class of AI systems with inherent adaptability and self-regularization capabilities.
	
	We have demonstrated the profound theoretical connection between this framework and fundamental physics, analyzed the geometric meaning of its expressiveness and regularization path, and detailed a practical computational method based on discrete differential geometry and automatic differentiation.
	
	This work is not an endpoint but the starting point of a grander journey. It lays the indispensable theoretical, computational, and conceptual foundation for ultimately realizing fully dynamic geometric-topological evolving meta-learners—intelligent agents capable of autonomously discovering and adapting to the most fundamental shapes of data. We foresee that this path towards ``learning geometry'' will not only spawn powerful new tools in AI but also deepen our understanding of the deep connections between information, geometry, and reality.
	
	\bibliographystyle{unsrt}
	\bibliography{references}
	
\end{document}